\documentclass{article}

\usepackage[T1]{fontenc}

\usepackage{PRIMEarxiv}

\usepackage[table,dvipsnames]{xcolor}
\usepackage{amsmath,amsfonts,amssymb}
\usepackage{dsfont}
\usepackage{graphicx}
\usepackage[colorlinks=true]{hyperref}
\hypersetup{
  linkcolor=black,
  citecolor=black,
  urlcolor=[RGB]{5,115,185}
}

\usepackage{gensymb}

\usepackage[sorting=none, backend=biber]{biblatex} %Imports biblatex package
\usepackage{tabularx}
\usepackage{multirow}
\usepackage{colortbl}
\usepackage{booktabs}
\usepackage{fancyhdr}

\newcolumntype{L}{>{\raggedright\arraybackslash}X}   
\newcolumntype{Y}{>{\centering\arraybackslash}X}
\newcolumntype{Z}{>{\raggedright\arraybackslash}p{3.5cm}}
\newcolumntype{A}[1]{>{\hspace*{-#1}\centering\arraybackslash}X}
\newcolumntype{B}{>{\centering\arraybackslash}p{5cm}}
\renewcommand{\arraystretch}{1.3}  % row height multiplier, default is 1

\definecolor{DarkBlue}{rgb}{0,0.28,0.67}

\newcommand{\sherpa}{\href{https://www.sherpa.ai/}{\textcolor{DarkBlue}{Sherpa.ai} }}

\fancypagestyle{firstpagestyle}{%
  \fancyhf{}%
  \fancyhead[R]{\includegraphics[scale=1.0]{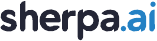}} % logo top-right
  \fancyfoot[C]{\thepage}%
}

\title{Federated Learning for Pediatric Pneumonia Detection: Enabling Collaborative Diagnosis Without Sharing Patient Data}
\author{
  {\LARGE \href{https://sherpa.ai/}{Sherpa.ai}} \\ \\
  research@sherpa.ai
}

\begin{document}

\maketitle
% %%%%%% PUT LOGO IN FIRST PAGE ONLY %%%%%%
\thispagestyle{firstpagestyle}
%%%%%%%%%%%%%%%%%%%%%%%%%%%%%%%%%%%%%%%%%%%

\begin{abstract}
Early and accurate pneumonia detection from chest X-rays (CXRs) is clinically critical to expedite treatment and isolation, reduce complications, and curb unnecessary antibiotic use. Although artificial intelligence (AI) substantially improves CXR-based detection, development is hindered by globally distributed data, high inter-hospital variability, and strict privacy regulations (e.g., HIPAA, GDPR) that make centralization impractical. These constraints are compounded by heterogeneous imaging protocols, uneven data availability, and the costs of transferring large medical images across geographically dispersed sites.

In this paper, we evaluate Federated Learning (FL) using the \sherpa FL platform, enabling multiple hospitals (nodes) to collaboratively train a CXR classifier for pneumonia while keeping data in place and private. Using the Pediatric Pneumonia Chest X-ray dataset, we simulate cross-hospital collaboration with non-independent and non-identically distributed (non-IID) data, reproducing real-world variability across institutions and jurisdictions. Our experiments demonstrate that collaborative and privacy-preserving training across multiple hospitals via FL led to a dramatic performance improvement — achieving 0.900 Accuracy and 0.966 ROC-AUC, corresponding to 47.5\% and 50.0\% gains over single-hospital models (0.610; 0.644), without transferring any patient CXR.

These results indicate that FL delivers high-performing, generalizable, secure and private pneumonia detection across healthcare networks, with data kept local. This is especially relevant for rare diseases, where FL enables secure multi-institutional collaboration without data movement, representing a breakthrough for accelerating diagnosis and treatment development in low-data domains.

\end{abstract}

\begin{figure}[h]
    \centering
    \includegraphics[width=0.60\linewidth]{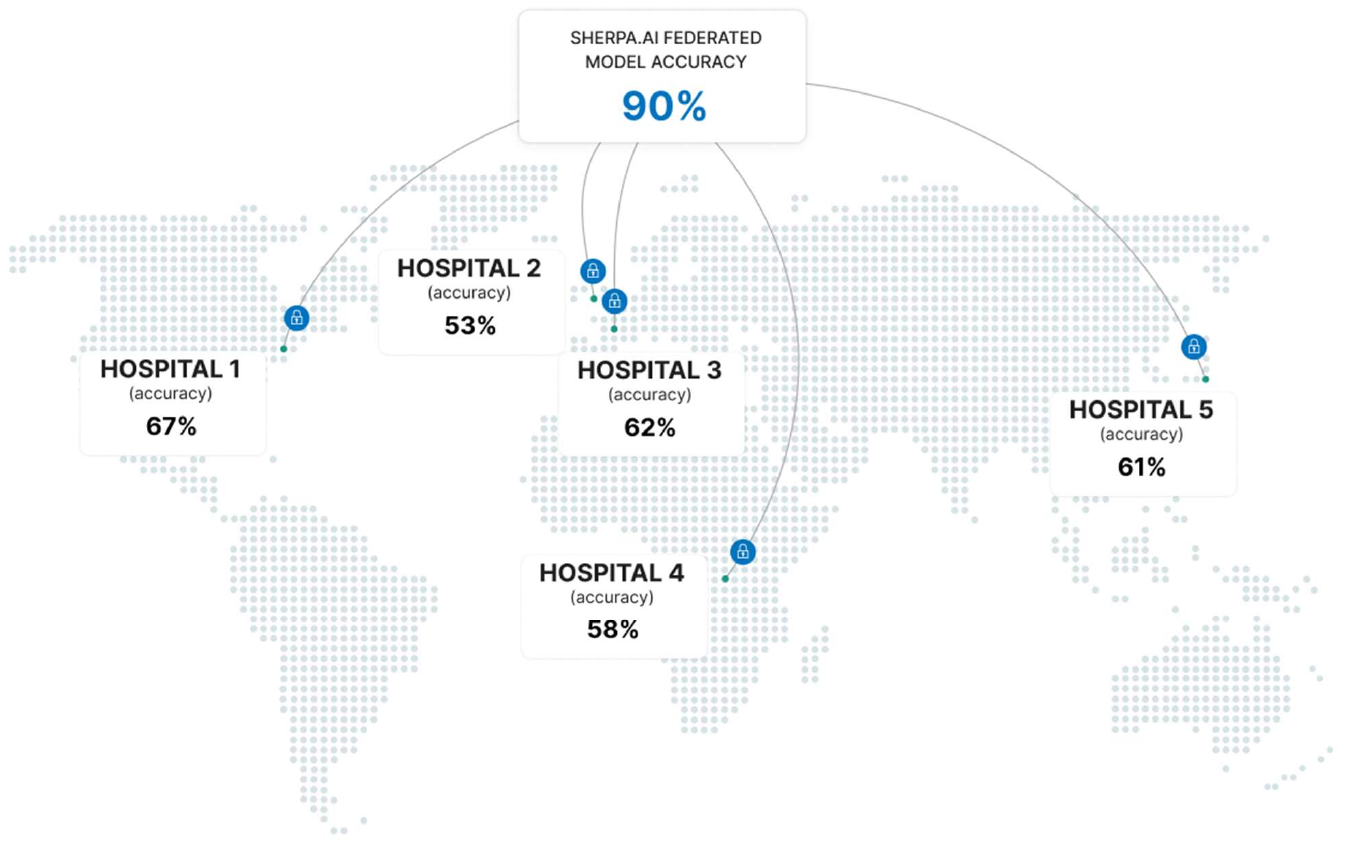}
    \caption{Global map illustrating hospitals joining the FL initiative to enhance model generalizability.}
    \label{fig:world_accuracy_comparison}
\end{figure}

\section{Introduction}

Pneumonia detection from chest X-ray (CXR) images is a critical task in medical image analysis, particularly for pediatric populations where the disease burden is high and early diagnosis is essential for reducing morbidity and mortality~\cite{rudan2013epidemiology}. Machine Learning (ML) automated methods for interpreting CXRs aim to assist radiologists by providing accurate, efficient, and scalable diagnostic tools, especially in low-resource settings where expert availability is limited.

Recent advances in deep learning have significantly improved performance in medical image classification, including pneumonia detection~\cite{rajpurkar2017chexnet, irvin2019chexpert}. Convolutional neural networks (CNNs) have demonstrated high diagnostic accuracy on publicly available datasets, such as the Pediatric Pneumonia Chest X-ray Dataset~\cite{kermany2018identifying}, and have shown potential for deployment in clinical workflows. These models typically rely on supervised learning using large, annotated datasets, and benefit from architectures originally developed for natural image tasks, such as ResNet, DenseNet, and EfficientNet~\cite{he2016deep, huang2017densely, tan2019efficientnet}.

Despite these advances, several challenges remain. First, the performance of deep models can degrade significantly when applied to data from institutions or populations different from the training set, due to domain shift and population heterogeneity~\cite{zech2018variable}. Second, medical image collections often have many more examples of one type of case than another. For instance, there may be thousands of X-rays from healthy children, but only a few from children with pneumonia. This imbalance makes it more challenging for an AI model to accurately recognize less common cases. Additionally, these datasets are typically small in scale due to the high cost of labeling and concerns about patient privacy. Third, the generalization of trained models across different hospitals, imaging protocols, and equipment remains limited, as differences in image quality, formats, and patient populations can render a model trained in one setting ineffective in another.

A key barrier to constructing robust, generalizable models in healthcare is the difficulty of data sharing. Due to stringent privacy regulations, such as HIPAA~\cite{hipaa1996} and GDPR~\cite{gdpr2016}, as well as institutional data governance policies, centralized aggregation of medical datasets is often infeasible. This restricts the ability to build diverse training sets that capture inter-institutional variability, thereby undermining model generalization to unseen clinical environments.

Federated Learning (FL)~\cite{mcmahan2017communication} provides a promising solution to this problem by enabling collaborative model training across multiple institutions without requiring raw data exchange (see Figure~\ref{fig:world_accuracy_comparison}). In particular, Horizontal FL (HFL)~\cite{yang2019federated}, where all participating sites (e.g., hospitals) share a common feature space but differ in sample (e.g., patients) distributions, is well-suited for medical imaging tasks involving consistent modalities like CXRs.

Applying HFL to pediatric pneumonia detection offers clear benefits but also practical challenges. It enables institutions to jointly train models that leverage broader data diversity without compromising patient privacy or violating regulatory constraints, thereby enhancing generalization across clinical settings and supporting continual learning as new data arrive. In practice, however, real-world federations rarely satisfy the independent and identically distributed (IID) assumption, as sites differ in case mix, imaging protocols, devices, and labeling practices, resulting in non-IID data. Such heterogeneity can induce client drift, unstable convergence, and bias toward dominant sites, degrading out-of-distribution generalization and fairness—issues documented in medical-imaging FL~\cite{xu2021federated}.

\section{Problem Formulation}

This section formalizes the task of pediatric pneumonia detection from CXR images in an FL setting. We first describe the medical image classification problem, with a focus on pediatric pneumonia, and then provide a formal definition of the supervised learning task, including its structure, objectives, and loss function formulation.

\subsection{Pneumonia Detection from CXRs}

Pneumonia detection is a fundamental task in medical image analysis, particularly for pediatric patients, where timely and accurate diagnosis is critical. CXR imaging is the standard diagnostic tool used to assess lung conditions in children. Automated systems aim to assist radiologists by classifying CXRs as either normal or pneumonia-positive, based on learned patterns indicative of infection, such as lung opacities. Unlike object detection, which involves spatial localization, this task is framed as image-level classification, where each image is associated with a single diagnostic label. Nevertheless, challenges such as subtle imaging features, overlapping anatomical structures, and inter-patient variability increase the complexity of the task.

These systems are typically deployed within large hospital networks and national health services, such as Mayo Clinic, Cleveland Clinic, Mount Sinai, Charit\'e--Universit\"atsmedizin Berlin, the U.K. National Health Service (NHS), and U.S. research agencies such as the National Institutes of Health (NIH), and must interoperate with widely used medical imaging and picture archiving and communication systems (PACS) from major vendors including Siemens Healthineers, GE Healthcare, Philips, Fujifilm, and Sectra.

\subsection{Related Work}

In the centralized paradigm, deep learning models, particularly CNNs, have achieved promising results in medical image classification tasks. Models such as CheXNet~\cite{rajpurkar2017chexnet}, DenseNet~\cite{huang2017densely}, and EfficientNet~\cite{tan2019efficientnet} have demonstrated high performance on thoracic disease classification using large-scale datasets like ChestX-ray8~\cite{wang2017chestx} and CheXpert~\cite{irvin2019chexpert}. For the detection of pediatric pneumonia specifically, Kermany et al.~\cite{kermany2018identifying} introduced a curated dataset of pediatric CXR images that has become a widely used benchmark. Singh et al.~\cite{singh2023pneumonia} propose a Quaternion Channel--Spatial Attention (QCSA) architecture that couples spatial/channel attention with quaternion residual backbones. Ayan et al.~\cite{ayan2022diagnosis} develop an ensemble of ImageNet-pretrained CNNs tailored to pediatric cohorts. Finally, Szepesi et al.~\cite{szepesi2022detection} present a deep CNN pipeline for pneumonia classification on chest radiographs. These works provide strong centralized references and architectural components (e.g., attention mechanisms, model ensembling) that complement FL and help contextualize cross-site results.

Recent studies on FL for pneumonia detection on CXRs increasingly focus on addressing cross-site non-IID data effects, label imbalance, and data scarcity while preserving privacy. Mabrouk et al.~\cite{mabrouk2023ensemble} introduce an ensemble FL approach that aggregates diverse client models to improve robustness under non-IID data, while Kareem et al.~\cite{kareem2023federated} present a general FL framework demonstrating end-to-end viability when data cannot be centralized. For pediatric cohorts, Pan et al.~\cite{pan2024efficient} emphasize communication- and computation-efficient protocols, and Biswas et al.~\cite{biswas2025flpnexainet} propose FLPneXAINet, which employs GAN-based augmentation to mitigate class imbalance and integrates XAI to enhance clinician trust.

Nevertheless, both centralized pipelines and current FL solutions face practical constraints in pediatric settings. Centralized state-of-the-art models typically assume access to large, curated, and centrally hosted data—an assumption often invalidated by privacy regulations, consent requirements, and institutional policies, which in turn induces a distribution shift at deployment. FL alleviates the data-sharing barrier by enabling cross-site training without moving raw data~\cite{kaissis2021end}; however, realizing consistent gains in practice still demands careful handling of cross-site heterogeneity (non-IID data, case mix, protocols, and devices), small cohort sizes, communication/orchestration overhead, and rigorous external validation and calibration across sites. Recent FL studies have begun to address these issues, but pediatric pneumonia remains especially challenging due to limited data availability and pronounced inter-site variability.

\subsection{Problem Description}
\label{sec:prob-description}

The pneumonia detection task is formulated as a binary classification problem for images. Given a CXR image \( x \in \mathcal{X} \subset \mathbb{R}^{H \times W} \), the goal is to predict a diagnostic label \( y \in \mathcal{Y} = \{0, 1\} \), where \( y = 0 \) denotes normal and \( y = 1 \) denotes pneumonia-positive.

\paragraph{Output Structure.}
The output is a single binary vector for each input image. Ground-truth labels are scalar values in \{0,1\}. Nevertheless, the classification task involves modeling complex imaging features and inter-institutional variability. The model must map each image to a posterior probability \( p_\theta(y \mid x) \), from which a hard label is derived via thresholding.

\paragraph{Objective.}
Given a federated dataset composed of \( K \) nodes (e.g., hospitals), each with a local dataset \( \mathcal{D}_k = \{(x_i^{(k)}, y_i^{(k)})\}_{i=1}^{N_k} \), where \( y_i^{(k)} \in \{0,1\} \) is a binary label, the global objective is to learn a model \( f_\theta: \mathcal{X} \to [0,1] \) that minimizes the expected binary cross-entropy loss over the combined (but decentralized) data distribution.

\paragraph{Loss Computation.}
The standard loss function for binary classification is the binary cross-entropy (BCEWithLogitsLoss):
\[
    \mathcal{L}_{\text{BCE}}(\theta) = - \frac{1}{N} \sum_{i=1}^{N} \Big[ y_i \log p_\theta(x_i) + (1 - y_i) \log \big(1 - p_\theta(x_i)\big) \Big],
\]
where \( y_i \in \{0, 1\} \) is the ground truth label, with \( 0 \) for negative and \( 1 \) for positive cases, and \( p_\theta(x_i) \) is the model’s predicted probability of the positive class given input \( x_i \). The loss is averaged over all training samples.

\section{Privacy-Preserving ML Solution}

This section outlines our privacy-preserving ML approach to medical image analysis, highlighting the motivation for FL, its regulatory implications, and the particularities of the HFL paradigm used in this work.

\subsection{Classical ML Approach}

In the standard centralized setting, medical image analysis is treated as a supervised learning problem. Given a labeled dataset $\mathcal{D} = \{(x_i, y_i)\}_{i=1}^N$, where each image $x_i \in \mathcal{X}$ (e.g., MRI, CT, histopathology) is paired with an annotation $y_i \in \mathcal{Y}$ (e.g., segmentation mask, classification label), the objective is to learn a function \( f_\theta: \mathcal{X} \to \mathcal{Y} \), parameterized by $\theta \in \mathbb{R}^d$, that generalizes to unseen patient data.

Model training proceeds by minimizing the empirical risk over $\mathcal{D}$:
\[
\min_{\theta \in \mathbb{R}^d} \mathcal{L}(\theta) = \frac{1}{N} \sum_{i=1}^N \ell(f_\theta(x_i), y_i),
\]
where $\ell: \mathcal{Y} \times \mathcal{Y} \rightarrow \mathbb{R}_{\geq 0}$ denotes the task-specific loss (e.g., cross-entropy for classification).

Optimization typically employs variants of stochastic gradient descent (SGD) or adaptive optimizers such as Adam:
\[
\theta_{t+1} = \theta_t - \eta_t \nabla_\theta \mathcal{L}(\theta_t),
\]
with learning rate $\eta_t$ at iteration $t$.

While effective, this paradigm assumes unrestricted access to centralized data, a premise that fails in many medical scenarios due to legal and ethical constraints on patient privacy, institutional data silos, and communication bottlenecks associated with large image volumes.

\subsection{Privacy and Regulatory Constraints}

Medical images are highly sensitive, often containing personally identifiable information (PII), and are protected under strict regulatory frameworks such as HIPAA (USA)~\cite{hipaa1996} and GDPR (EU)~\cite{gdpr2016}. These frameworks prohibit unrestricted data transfer, especially when involving third parties or cloud servers.

Even with anonymization, transmitting large datasets (e.g., whole-slide images or 3D MRI scans) is bandwidth-intensive and logistically complex. These factors hinder the feasibility of centralized training for many institutions, especially in regions with limited digital infrastructure.

A more secure and efficient alternative is to transmit only model parameters or updates, rather than the raw images, thus enabling collaborative learning without data exposure. FL provides such a mechanism by allowing institutions to co-train models locally and aggregate knowledge globally, all while adhering to privacy mandates. Furthermore, this approach reduces transmission overhead and contributes to the goals of energy-efficient AI~\cite{mehta2023review}.

\subsection{Introduction to FL}
\label{sec:fedlearning-overview}

FL~\cite{mcmahan2017communication} enables decentralized training by distributing the model to nodes, each of which trains locally on private data and periodically shares updates with a coordinating server. The server aggregates the updates to form a new global model, which is redistributed to the nodes for continued training.

This process maintains data locality and enhances privacy, but presents new challenges in optimization, particularly under heterogeneous (non-IID) data distributions. Such settings are common in medical contexts, where datasets differ in terms of demographics, imaging hardware, acquisition protocols, or institutional practices~\cite{lu2024federated}.

FL has shown promise in medical imaging tasks, including tumor segmentation~\cite{sheller2020federated} and disease classification~\cite{li2020federated1, feki2021federated}. However, many of these studies assume idealized data homogeneity or static node participation. 

\subsubsection{FL Paradigms}
\label{sec:fedlearning-paradigms}

FL encompasses different paradigms depending on how data is distributed across nodes:
\begin{itemize}
    \item \textbf{HFL}: Nodes share the same feature space (e.g., all hospitals collect CXRs), but have disjoint patient cohorts. Each node holds different samples drawn from the same modality and label space.
    
    \item \textbf{Vertical FL (VFL)}: Nodes hold different feature subsets for the same patient set (e.g., imaging features at one institution, genomics at another), and collaboration requires secure entity alignment.
\end{itemize}

This work exclusively considers HFL, reflecting realistic hospital collaborations where institutions collect data from similar imaging pipelines but on different patient populations.

\subsubsection{HFL}
\label{sec:fedlearning-hfl}

In the HFL setting, each node \(k\) owns a private dataset:
\[
\mathcal{D}_k = \big\{ (x_i^k, y_i^k) \big\}_{i=1}^{N_k},
\]
where all inputs \(x_i^k \in \mathbb{R}^p\) share the same feature structure, but the data distribution and sample size \(N_k\) may vary significantly across nodes.

Training proceeds in the following stages:
\begin{enumerate}
    \item \textbf{Local update:} Each node optimizes its local loss:
    \[
    \mathcal{L}_k(\theta) = \frac{1}{N_k} \sum_{i=1}^{N_k} \ell(f_\theta(x_i^k), y_i^k),
    \]
    and computes updated parameters \(\theta_k\).
    
    \item \textbf{Aggregation:} The server collects the local models \(\{\theta_k\}_{k=1}^K\) and forms a new global model, typically via weighted averaging (e.g., FedAvg).
    
    \item \textbf{Broadcast:} The updated global parameters \(\theta\) are sent back to each node.
\end{enumerate}

This iterative procedure continues until the model converges or satisfies the stopping criteria. HFL offers a principled solution for collaborative training in medical imaging, leveraging the diversity of datasets across hospitals while maintaining patient confidentiality.

\section{Centralized Dataset and Preprocessing}

In this section, we describe the dataset, outline the preprocessing steps required for medical imaging analysis, and present the centralized deep learning architecture used as a baseline.

\subsection{Description of the Dataset}

The dataset used in this study is the Pediatric Pneumonia Chest X-ray Dataset, originally curated by Kermany et al.~\cite{kermany2018identifying} and made publicly available through the Guangzhou Women and Children's Medical Center. It consists of pediatric chest radiographs from children aged 1 to 5 years, including both healthy and pneumonia-positive cases. The data was collected in a clinical setting and manually labeled by certified radiologists.

\begin{figure}[h!]
\centering
\includegraphics[width=0.3\linewidth]{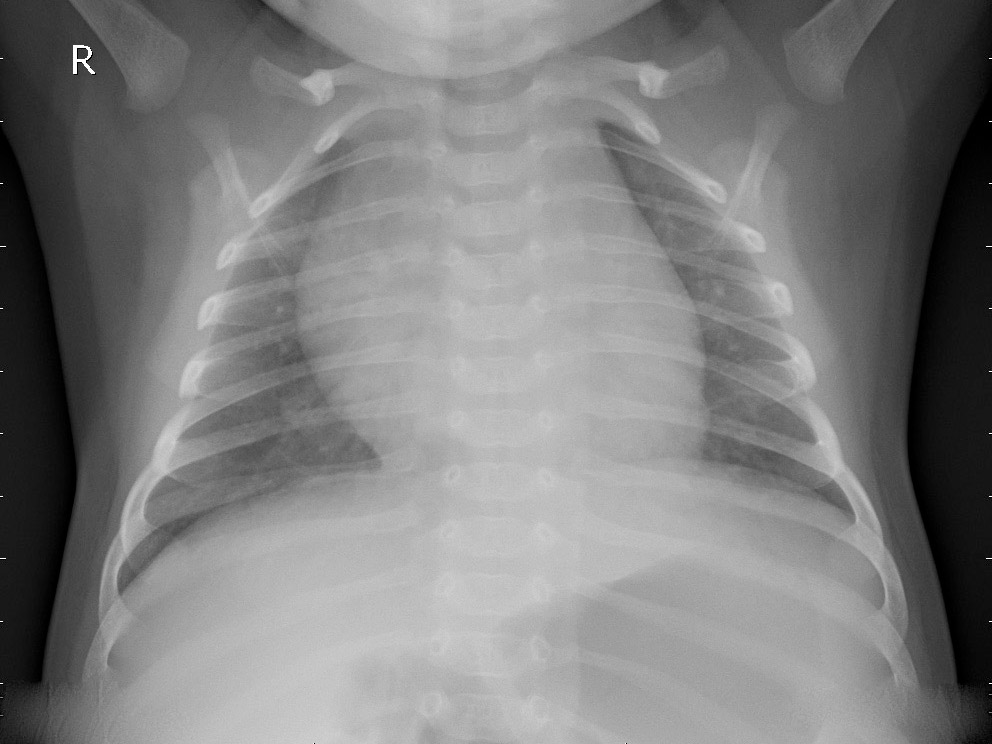}
\includegraphics[width=0.3\linewidth]{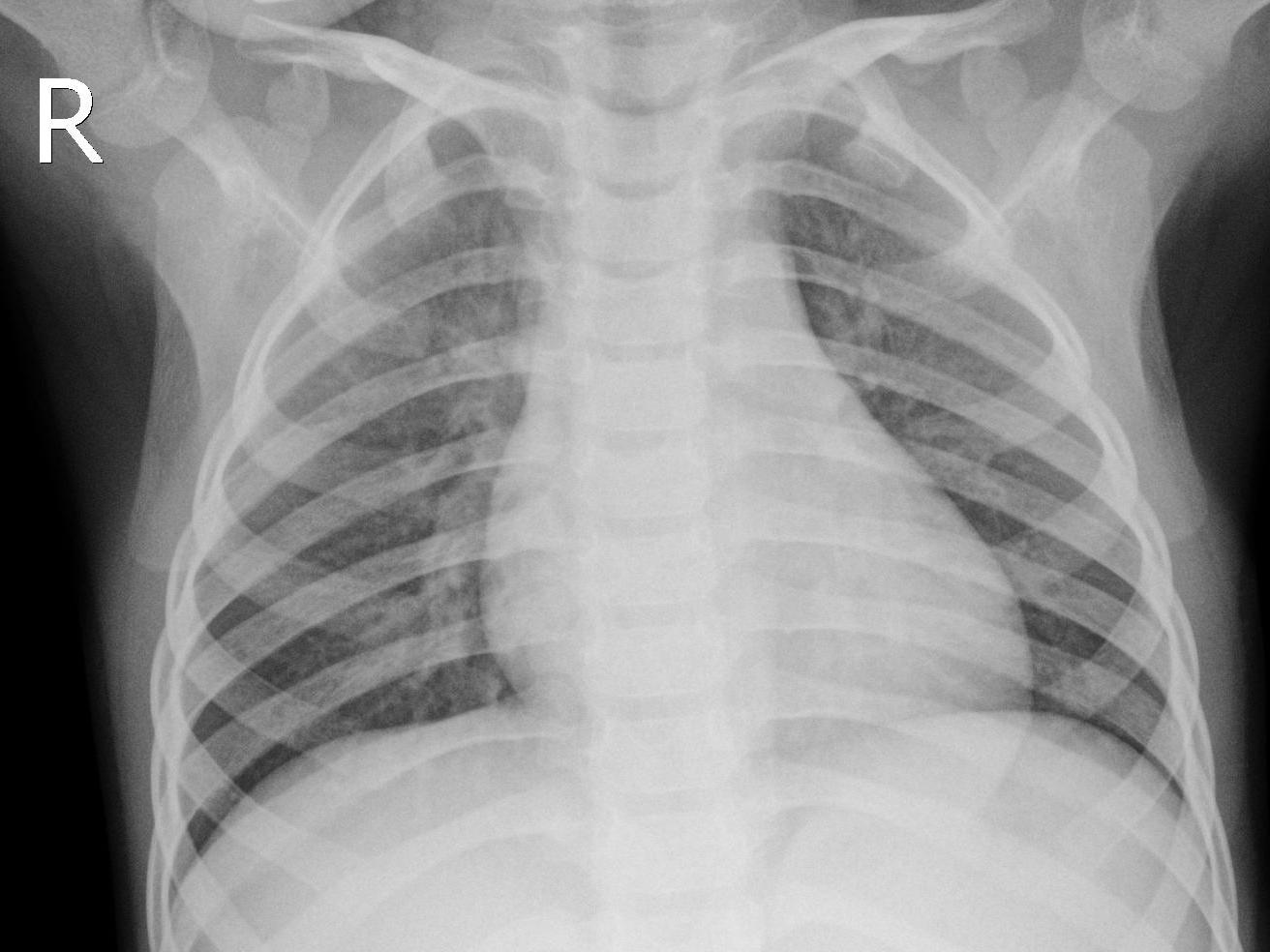}
\includegraphics[width=0.3\linewidth]{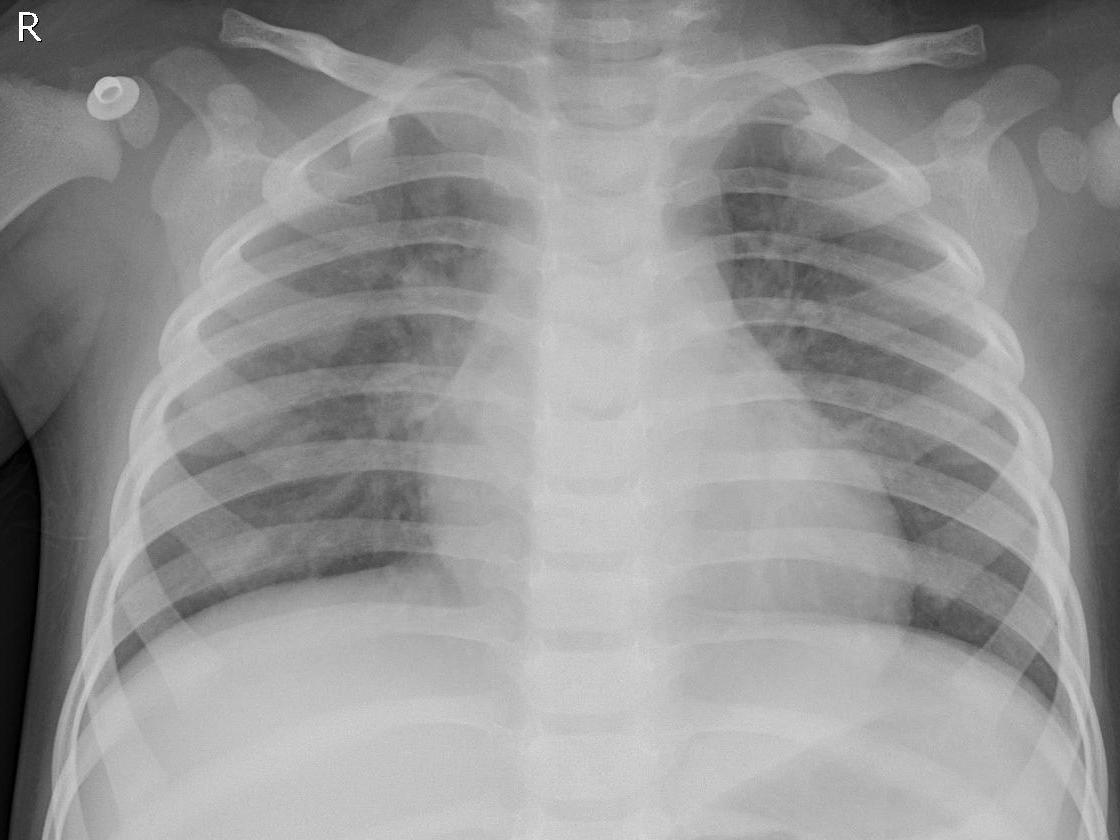}
\caption{Example pediatric CXR images: (left) normal, (center) viral pneumonia, (right) bacterial pneumonia.}
\label{fig:example-chest-xray}
\end{figure}

This dataset comprises 5,856 CXR images, divided into three categories: \texttt{Normal}, \texttt{Bacterial Pneumonia}, and \texttt{Viral Pneumonia}. For the purpose of this study, we frame the problem as a two-class classification task distinguishing between a healthy person (\texttt{Normal}) and one having any type of pneumonia (\texttt{Pneumonia}). This choice aligns with previous studies in this field~\cite{singh2023pneumonia,ayan2022diagnosis,szepesi2022detection}. The sample images from each class are shown in Figure~\ref{fig:example-chest-xray}. 

The original dataset is divided into training (5,210 images), validation (originally, the subset did not contain all classes, so we extended it to 150 images), and testing (496 images).

The main hyperparameters and settings are as follows:

\begin{itemize}
\item \textbf{Input image size:} Varies.
\item \textbf{Optimizer:} Adam.
\item \textbf{Batch size:} 32.
\item \textbf{Epochs:} 25.
\item \textbf{Learning rate:} $5\cdot 10^{-5}$.
\item \textbf{Loss function:} Binary cross-entropy.
\end{itemize}

These hyperparameters were obtained empirically using the proprietary \sherpa FL platform as a means of experimentation, using different hyperparameter ranges.

\subsection{Preprocessing of the Dataset}

Given that CXR images in the dataset are grayscale with varying resolutions, preprocessing is essential to ensure uniformity, improve model generalization, and adapt to the input format of modern convolutional architectures. The preprocessing pipeline applied in our implementation includes the following steps:

\begin{itemize}
\item \textit{Resizing:} All images are resized to \(224 \times 224\) pixels to ensure consistency with standard backbone input dimensions.
\item \textit{Data Augmentation:} To mitigate overfitting and enhance generalization, we employ a diverse augmentation pipeline using the \texttt{Albumentations} library. With an overall probability of 0.75, the following transformations are applied:
\begin{itemize}
    \item \texttt{HorizontalFlip} with probability 0.5
    \item \texttt{Affine} transformations: random scaling ($\pm$15\%), rotation ($\pm$30\degree)
    \item \texttt{RandomGamma}, \texttt{RandomBrightnessContrast}
    \item \texttt{GaussNoise} with standard deviation up to 0.05
    \item \texttt{Blur} or \texttt{MedianBlur}, \texttt{OpticalDistortion}, \texttt{GridDistortion}, \texttt{ElasticTransform}
    \item \texttt{RandomFog}, \texttt{CLAHE}
\end{itemize}
All augmentations are wrapped in a probabilistic \texttt{Compose} block, applied only to the training set. The validation set is resized but not augmented.
\item \textit{Normalization:} After augmentation, images are normalized using ImageNet statistics: mean \( = [0.485, 0.456, 0.406] \), standard deviation \( = [0.229, 0.224, 0.225] \), applied per channel.
\item \textit{Tensor Conversion:} The processed images are converted to PyTorch-compatible \texttt{tensor} objects using \texttt{ToTensorV2()}.
\item \textit{Label Encoding:} Class labels are converted to integer indices.
\end{itemize}

\subsection{Centralized Architecture}

\begin{figure}[ht]
    \centering
    \includegraphics[scale=0.7]{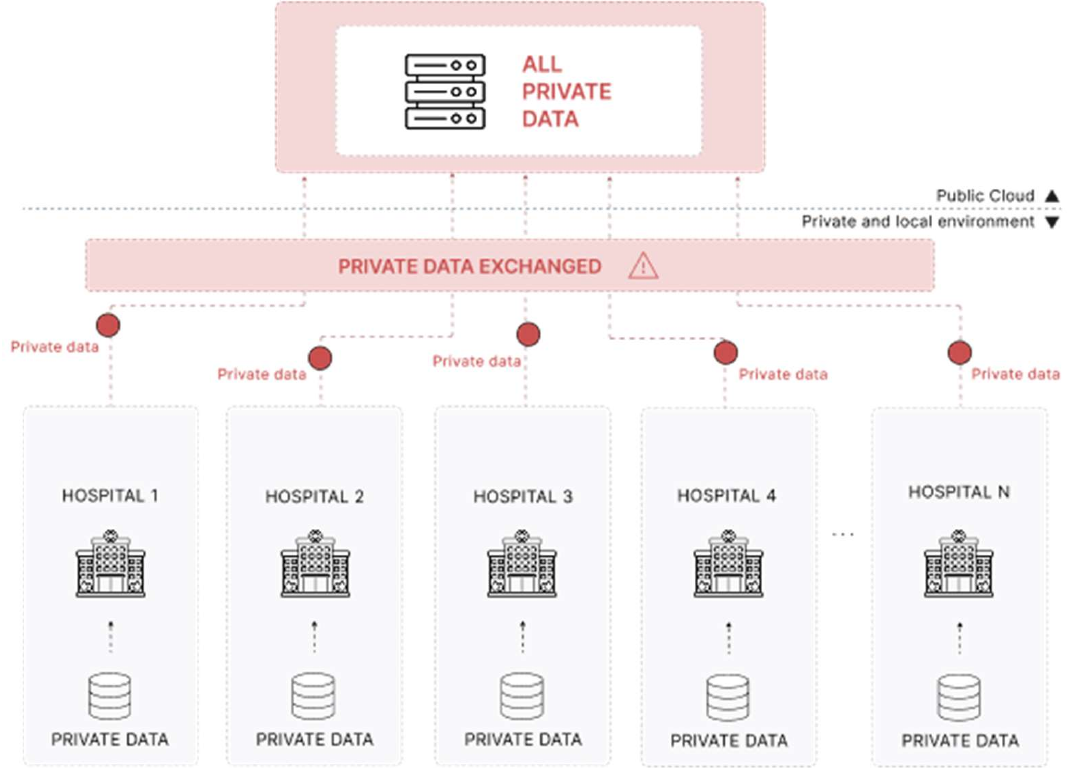}
    \caption{Proposed architecture for centralized training.}
    \label{fig: cen-arch}
\end{figure}

As shown in Figure~\ref{fig: cen-arch}, in a centralized training scenario, all participating hospitals would first need to pool their patient CXR data into a single centralized server. This would involve transferring locally stored data from each institution to a common data repository, enabling the model to be trained on the entire dataset as a unified whole. Such an approach assumes that cross-institutional data sharing is both legally permissible and ethically acceptable, which is often not the case in medical settings due to privacy laws, data governance restrictions, and institutional policies. While centralized training can potentially yield optimal performance due to full data availability, it disregards the practical constraints of real-world clinical environments where data silos are prevalent and privacy preservation is paramount. Nevertheless, to establish a performance baseline for the federated setting, we first train a centralized CNN on the entire dataset. Our suggested architecture follows a standard image classification pipeline using a ResNet18~\cite{he2016deep} backbone pretrained on ImageNet~\cite{deng2009imagenet}.

The classifier head replaces the original fully connected layer of ResNet18 with a custom classification stack:

\begin{itemize}
\item Linear layer with 512 output units
\item Batch normalization
\item ReLU activation
\item Dropout (rate = 0.5)
\item Final linear layer with \( c = 1 \) output output unit (single logit)
\end{itemize}

During evaluation, class probabilities are computed by applying a sigmoid activation function to the final logit.

Training was conducted using PyTorch. The loss function is a binary cross-entropy. The model is optimized using the Adam optimizer.

The centralized model typically converges in fewer than 25 epochs and serves as the performance benchmark for the FL experiments.

\section{Proposed Privacy-Preserving Solution through FL}

In this section, we describe the FL architecture, node creation strategy, and evaluation metrics used for performance assessment under HFL.

\subsection{FL Architecture}

\begin{figure}[ht]
    \centering
    \includegraphics[scale=0.7]{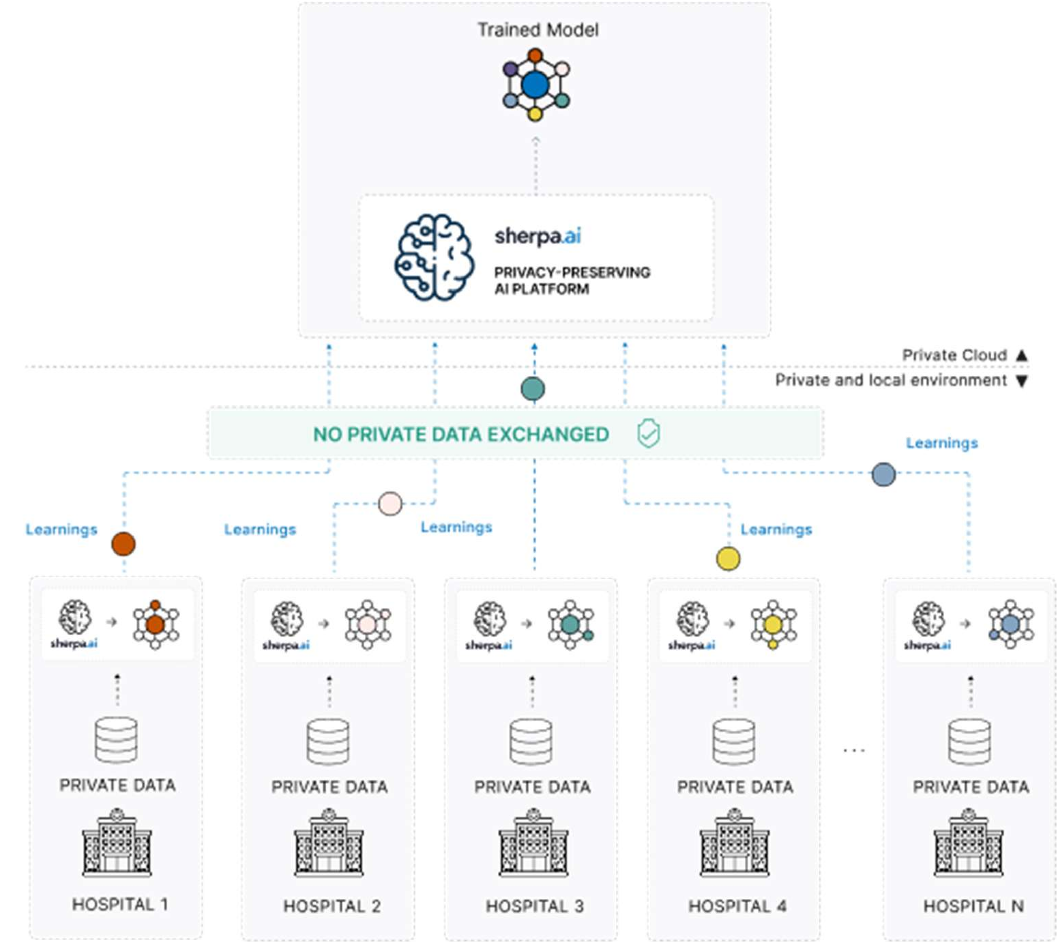}
    \caption{Proposed architecture for federated training with privacy-preserving orchestration by \sherpa.}
    \label{fig: fl-arch}
\end{figure}

We implement the HFL pipeline using the \sherpa FL platform, as illustrated in Figure~\ref{fig: fl-arch}. A global model is initialized and transmitted to all participating nodes. Each node trains the model locally using its private subset of CXR images. After a specified number of local epochs, nodes send their model weights to the server, which performs aggregation using the Federated Averaging (FedAvg) algorithm~\cite{mcmahan2017communication}. The updated global model is then redistributed to all nodes. This process is repeated for multiple rounds until convergence.

In contrast to the centralized setting, no raw images are exchanged across nodes. This ensures compliance with medical data privacy regulations, such as GDPR and HIPAA, by keeping all sensitive health information on-device or within the data-holding institution.

\subsection{Creation of Nodes}
\label{sec:nodes_creation}

To simulate heterogeneous data distributions across federated nodes, we employed a Dirichlet-based partitioning strategy using the FedArtML tool~\cite{gutierrez2024fedartml, li2022federated}. Given a centralized dataset $\mathcal{D} = \{(x_i, y_i)\}_{i=1}^N$ with binary labels $y_i \in \{0,1\}$ representing \texttt{Normal} and \texttt{Pneumonia} classes, we partitioned the data among $K=5$ federated nodes by sampling class proportions from a Dirichlet distribution $\text{Dir}(\alpha \mathbf{1}_C)$, where $\alpha$ is the concentration parameter and $C=2$ is the number of classes. For each node $k$, the proportion vector $\mathbf{p}_k = (p_{k,0}, p_{k,1}) \sim \text{Dir}(\alpha, \alpha)$ determines the relative class distribution, with smaller $\alpha$ values inducing higher statistical heterogeneity across nodes. For our experiments, we selected $\alpha = 0.25$, which generates a highly non-IID data scenario. Specifically, we used the official partitions provided for the training and validation folders (corresponding to 89\% and 2.5\% of the total), and each subset was independently partitioned across the \( K = 5 \) nodes using FedArtML to preserve the original data splits while inducing controlled non-IID distributions within each subset.

The resulting federated datasets $\{\mathcal{D}_k\}_{k=1}^K$ exhibit label skewness quantified by the Hellinger distance~\cite{goussakov2020hellinger}:
\[
H\left(\mathbf{p}_k, \mathbf{p}_j\right) = \frac{1}{\sqrt{2}}\left\| \sqrt{\mathbf{p}_k} - \sqrt{\mathbf{p}_j}\right \|_2
\]
between node distributions. This approach ensures systematic control over the degree of non-IID data while maintaining the original data integrity through file-based partitioning of the CXR imaging dataset. An example of such a partition can be found in Table~\ref{tab:federated_data_split}. It simulates non-IID real-world hospital distributions.

\begin{table}[htbp]
\scriptsize
\setlength{\tabcolsep}{4pt}             
  \renewcommand{\arraystretch}{1.2}
  \centering
  \begin{tabularx}{0.75\textwidth}{L|Y|Y|Y|Y}      
    \toprule
    \textbf{Node} & \textbf{Normal} & \textbf{Pneumonia} & \textbf{Total} & \textbf{Normal Ratio} \\
    \midrule
     0 & 38 & 1482 & 1520 & 0.025 \\
     1 & 53 & 1 & 54 & 0.981 \\
     2 & 197 & 2363 & 2560 & 0.077 \\
     3 & 1033 & 1 & 1034 & 0.999 \\
     4 & 20 & 22 & 42 & 0.476 \\
    \midrule
    \textbf{Total} & \textbf{1341} & \textbf{3869} & \textbf{5210} & \textbf{0.257} \\
    \bottomrule
  \end{tabularx}
  \vspace{5mm}
  \caption{Train data distribution across federated nodes, generated by partitioning with a Dirichlet distribution with $\alpha = 0.25$ ($H\left(\mathbf{p}_k, \mathbf{p}_j\right) = 0.601$). The normal ratio indicates the proportion of Normal class samples in each node, illustrating the class and cardinality imbalance characteristic of federated medical imaging scenarios.}
  \label{tab:federated_data_split}
\end{table}

Each node retains exclusive access to its local data and trains its local model copy independently. Local training is conducted for one epoch per communication round, followed by global model synchronization via a central aggregator. This setup mimics real-world deployments where hospitals or medical centers collaborate under privacy constraints. The FL framework thus enables secure, decentralized training while respecting data locality, legal regulations, and institutional autonomy.

\section{Experiment}

We conducted a series of experiments to evaluate the performance of centralized, local, and FL strategies for pediatric pneumonia detection from CXRs inspired by the methodology of~\cite{kaissis2021end}. The models were implemented using PyTorch, with a ResNet18 backbone pretrained on ImageNet. Each model was trained under three settings:

\begin{itemize}
\item \textbf{Centralized:} A baseline model trained on the entire training dataset, assuming full data sharing across institutions.
\item \textbf{Single-Hospital:} (Local Baseline) Independent models trained on each node's local dataset (hospital) without any inter-node communication.
\item \textbf{Federated:} A global model trained using HFL across the \( K = 5 \) nodes. Model updates were aggregated at each communication round using a central server, without exposing local data.
\end{itemize}

To ensure statistical robustness and evaluate the variability introduced by different data allocations, we repeat each experimental configuration 10 times across independent runs, using different random seeds. In each run, a new set of five node-specific datasets is generated, resulting in 10 distinct federated data configurations.

All models, regardless of the training setting, were evaluated on a unified test set using the official test folder from the dataset, which corresponds to 8.5\% of the total data. This centralized test set was held out during training and provides a representative, consistent evaluation of generalization across the full data distribution.

\subsection{Evaluation Metrics}
\label{sec:metrics}

To assess the performance of the proposed models in the pediatric pneumonia detection task, we employ a suite of standard binary classification metrics: Precision, Recall, F1-score, and Area Under the Receiver Operating Characteristic Curve (ROC-AUC). All metrics are computed with respect to the positive class.

\begin{itemize}
    \item \textbf{Accuracy} measures the proportion of correctly classified samples among all test samples. While commonly used, it can be insufficient in cases of class imbalance.
    \begin{equation}
        \text{Accuracy} = \frac{1}{N} \sum_{i=1}^{N} \mathds{1}(\hat{y}_i = y_i)
    \end{equation}

    \item \textbf{Precision} quantifies the proportion of correctly predicted positive samples out of all samples predicted as positive.
    \begin{equation}
        \text{Precision} = \frac{\text{TP}}{\text{TP} + \text{FP}}
    \end{equation}

    \item \textbf{Recall} (or sensitivity) measures the proportion of correctly predicted positive samples out of all positive samples.
    \begin{equation}
        \text{Recall} = \frac{\text{TP}}{\text{TP} + \text{FN}}
    \end{equation}

    \item \textbf{F1-score} is the harmonic mean of Precision and Recall, providing a balance between false positives and false negatives.
    \begin{equation}
        \text{F1} = 2 \cdot \frac{\text{Precision} \cdot \text{Recall}}{\text{Precision} + \text{Recall}}
    \end{equation}

    \item \textbf{ROC-AUC} evaluates the trade-off between the true positive rate and false positive rate across thresholds. For binary classification, we report the single AUC value corresponding to the positive class.
\end{itemize}

\subsection{Experimentation testbed}

All experiments were conducted using a machine equipped with 1TB of disk space, an Intel Core i7-13700K processor at 3.4 GHz, 96 GB of RAM, an NVIDIA RTX A4000 GPU with 16 GB of RAM, a Ubuntu 24.04 operating system, and Python 3.11.

\subsection{Results}

Table~\ref{tab:performance_comparison} summarizes the overall classification metrics for models trained under three paradigms: local training at each node, FL across all nodes, and centralized training.

\begin{table}[htbp]
\scriptsize
\setlength{\tabcolsep}{4pt}
\renewcommand{\arraystretch}{1.2}
\centering
\begin{tabularx}{0.75\textwidth}{L|Y|Y|Y}
  \toprule
  \textbf{Metric} & \textbf{Single-Hospital} & \textbf{Centralized} & \textbf{Federated (Sherpa.ai)} \\
  \midrule
  Accuracy (STD)    & 0.610 (0.052) & 0.894 (0.012) & 0.900 (0.026) \\
  Precision (STD)    & 0.587 (0.130) & 0.882 (0.024) & 0.936 (0.034) \\
  Recall (STD)       & 0.663 (0.137) & 0.957 (0.036) & 0.903 (0.062) \\
  F1-score (STD)     & 0.580 (0.106) & 0.917 (0.010) & 0.917 (0.026) \\
  ROC-AUC (STD)      & 0.644 (0.085) & 0.958 (0.007) & 0.966 (0.006) \\
  \bottomrule
\end{tabularx}
\vspace{5mm}
\caption{Performance metrics for the Single-hospital, centralized, and federated settings. Results are reported as mean (standard deviation - STD) over 10 random trials, evaluated on a unified test set.}
\label{tab:performance_comparison}
\end{table}

\begin{figure}[htbp]
    \centering
    \includegraphics[width=.6\linewidth]{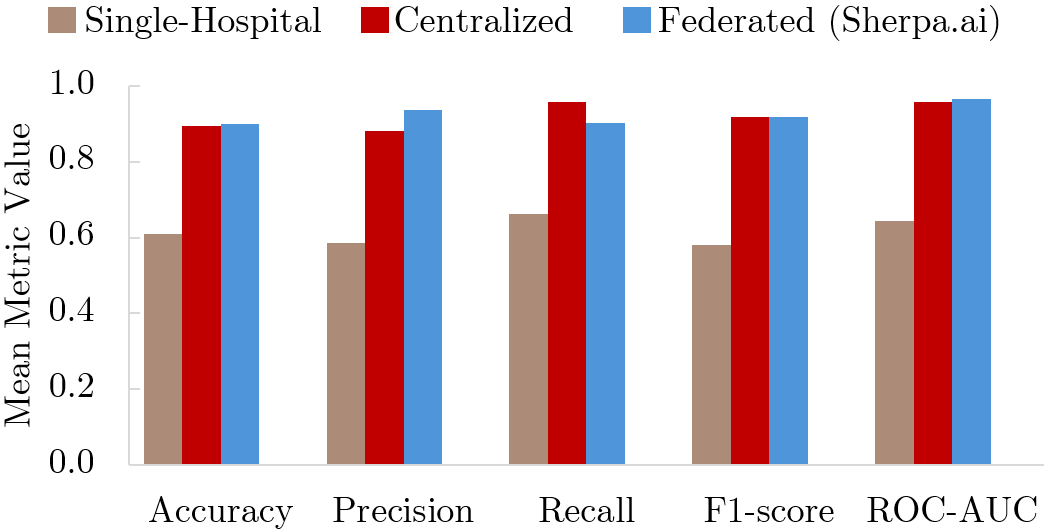}
    \caption{Mean metric values comparing Single-Hospital, Centralized, and Federated training settings.}
    \label{fig:res}
\end{figure}

% \noindent
As shown in Table~\ref{tab:performance_comparison} and Figure~\ref{fig:res}, models trained locally on each node's data exhibit moderate but highly variable performance, reflecting the limitations imposed by isolated, data-constrained environments. In contrast, the federated model achieves strong performance across all metrics. With an Accuracy of 0.900 and an F1-score of 0.917, it substantially outperforms the Single-Hospital, and approaches the performance of the centralized model. The federated model attains an ROC-AUC of 0.966, closely matching the centralized model (0.958) while significantly outperforming the Single-Hospital setting (0.644). In terms of precision and recall, the federated approach again demonstrates balanced performance with 0.936 and 0.903, respectively, bridging the gap between local and centralized training paradigms. In particular, the results are consistent with previous work~\cite{singh2023pneumonia,ayan2022diagnosis,szepesi2022detection}, particularly in terms of the ROC-AUC (a key performance metric for pneumonia prediction), where our values closely match those reported in the literature.

These results demonstrate the federated model’s capacity to generalize across data distributions, achieving performance that closely parallels that of the centralized model, despite the lack of direct data sharing. Furthermore, the narrow standard deviations across all metrics in the federated setting indicate greater stability and robustness over the 10 randomized trials, reinforcing the reliability of the approach.

\section{Conclusions}
\label{sec:Conclusions}

% Our experiments demonstrate that using the \sherpa FL platform, multiple institutions can collaboratively improve the performance for pediatric pneumonia models using CXR while keeping data local and private. Across ten randomized non-IID partitions, the federated model reliably surpassed Single-Hospital baselines and closely matched centralized performance in terms of Accuracy, F1-score, and ROC-AUC.

% Our findings demonstrate that FL facilitates privacy-preserving and regulation-compliant collaboration, making it suitable for clinical integration. Hospitals can seamlessly deploy plug-and-play, site-specific nodes within existing infrastructures, incorporating clinician-in-the-loop evaluation.

Our experiments demonstrate that, using the \sherpa FL platform, multiple institutions can collaboratively improve the performance of pediatric pneumonia detection models from CXR while keeping all data local and private. Across ten randomized non-IID partitions, the federated model consistently outperformed single-hospital baselines and closely matched centralized performance in terms of Accuracy, F1-score, and ROC-AUC.

These findings confirm that FL enables privacy-preserving, regulation-compliant collaboration among hospitals and research centers — even when institutions are located in different geographic regions governed by distinct legal and ethical frameworks such as HIPAA, GDPR, and other national data-protection laws. The ability to jointly train models without moving sensitive data ensures both compliance and scalability, providing a viable path for cross-border medical AI collaboration.

Beyond pediatric pneumonia, this approach is especially significant for rare diseases, where patient data are extremely limited and dispersed globally. By securely combining knowledge from distributed cohorts, FL could represent a major breakthrough in accelerating diagnostic accuracy, biomarker discovery, and treatment development in low-data clinical domains. Hospitals can seamlessly deploy plug-and-play, site-specific nodes within existing infrastructures, integrating clinician-in-the-loop validation while ensuring patient privacy and data sovereignty.

\section{Future Work}

In future work, we aim to extend this study from HFL to VFL — a setting in which multiple institutions contribute complementary modalities related to the same patients. For instance, while one hospital might provide chest X-ray images, another could contribute electronic health records, blood analyses, voice samples, or clinical metadata, together enabling richer multimodal learning without sharing patients' data. Such an extension could substantially enhance diagnostic performance, allowing models to integrate cross-domain biological and clinical signals across institutions while maintaining full privacy.
 
\sherpa has pioneered research in this direction through a novel paradigm named Blind Vertical Federated Learning~\cite{acero2025sherpa}, which introduces a new level of privacy preservation and computational efficiency. This approach achieves faster convergence and offers stronger confidentiality guarantees than conventional aggregation methods, pointing toward a new generation of secure and scalable collaborative intelligence in healthcare.

% Future work should explore scaling to larger numbers of nodes (to better reflect healthcare networks), improving communication efficiency, and incorporating more advanced aggregation strategies such as FedProx~\cite{li2020federated} and adaptive federated optimization methods. This could include systematic scalability studies across increasing node counts, profiling bandwidth/latency trade-offs for communication-efficient training, and ablation analyses contrasting FedProx with adaptive optimizers under varying scenarios.

% \newpage
% \section*{Appendix}

% \renewcommand{\thesection}{\Alph{section}} % sections labeled A, B, C...
% \setcounter{section}{0} % reset counter so first one starts at A

\section*{Contributions and Acknowledgments} \label{app:A}

% Alex Acero

Daniel M. Jimenez-Gutierrez

Enrique Zuazua

Joaquin Del Rio

Oleksii Sliusarenko

Xabi Uribe-Etxebarria

\vspace{5mm}
The authors are presented in alphabetical order by first name.

%%%%%%%%%%%%%%%%%%%%%%%%%%%%%%%%%%%%%%%%%%%%%%%%%%%%

%%%%%%%%%%%%%%%%%%%%%%%%%%%%%%%%%%%%%%%%%%%%%%%%%%%%
%%% Bibliography
%%%%%%%%%%%%%%%%%%%%%%%%%%%%%%%%%%%%%%%%%%%%%%%%%%%%
\printbibliography %Prints bibliography

\end{document}